\documentclass[conference]{IEEEtran}
\bstctlcite{IEEEexample:BSTcontrol}
\ifCLASSINFOpdf

\else

\fi

\usepackage{graphicx}
\usepackage{subcaption}
\usepackage{float}
\usepackage{url}
\usepackage{xcolor}
\usepackage{hyperref}

\usepackage{amsmath,amssymb,epsfig}
\newcommand{\ds}{\displaystyle}
\def\bomega{{\boldsymbol{\omega}}}

\begin{document}
\title{AFP-SRC: Identification of Antifreeze Proteins Using Sparse Representation Classifier}
\author{\IEEEauthorblockN{Shujaat Khan\IEEEauthorrefmark{1}, Muhammad Usman\IEEEauthorrefmark{2}, Abdul Wahab\IEEEauthorrefmark{3}}
    \IEEEauthorblockA{\\
        \IEEEauthorrefmark{1}Department of Bio and Brain Engineering, \\
        Korea Advanced Institute of Science and Technology (KAIST), Daejeon 34141, Republic of Korea.\\
        Email: shujaat@kaist.ac.kr\\
        \IEEEauthorrefmark{2}Department of Computer Engineering, \\
        College of IT Convergence Engineering, Chosun University, Gwangju 61452, Republic of Korea.\\
        Email: usman@chousn.kr \\
        \IEEEauthorrefmark{3}Department of Mathematics, \\ School of Sciences and Humanities, Nazarbayev University,  53 Kabanbay Batyr Avenue, Nur-Sultan 010000, Kazakhstan. \\ Email: abdul.wahab@nu.edu.kz
    }
}
\maketitle
\footnote{``This preprint has not undergone any post-submission improvements or corrections. The Version of Record of this article is published in Neural Computing and Applications, and is available online at https://doi.org/10.1007/s00521-021-06558-7''.}	
\begin{abstract}
Species living in the extreme cold environment fight against the harsh conditions using antifreeze proteins (AFPs), that manipulates the freezing mechanism of water in more than one way. This amazing nature of AFP turns out to be extremely useful in several industrial and medical applications. The lack of similarity in their structure and sequence makes their prediction an arduous task and identifying them experimentally in the wet-lab is time-consuming and expensive. In this research, we propose a computational framework for the prediction of AFPs which is essentially based on a sample-specific classification method using the sparse reconstruction. A linear model and an over-complete dictionary matrix of known AFPs are used to predict a sparse class-label vector that provides a sample-association score. Delta-rule is applied for the reconstruction of two pseudo-samples using lower and upper parts of the sample-association vector and based on the minimum recovery score, class labels are assigned. We compare our approach with contemporary methods on a standard dataset and the proposed method is found to outperform in terms of Balanced accuracy and Youden's index. The MATLAB implementation of the proposed method is available at the author's GitHub page (\href{https://github.com/Shujaat123/AFP-SRC}{https://github.com/Shujaat123/AFP-SRC}).

\end{abstract}
\begin{IEEEkeywords}
	\normalfont{Over-complete dictionary, basis-pursuit, sample specific classification, antifreeze proteins (AFPs), amino acid composition (AAC), di-peptide composition (DPC),  sparse reconstruction classification (SRC).}
\end{IEEEkeywords}
	
\IEEEpeerreviewmaketitle

\section{Introduction}
Antifreeze proteins (AFPs) are essential for the species living in an extremely cold environment. They protect them from freezing by manipulating the freezing mechanism of water through thermal hysteresis. AFPs are widely used in several industrial and medical applications such as cryo-preservation and food products \cite{prathalingam2006impact,qadeer2015efficiency,naing2019brief,griffith1995antifreeze,duman2014use,rubinsky1994freezing}.  Chemically, they appear in a variety of structures and have little sequence and structural similarity which makes their search a challenging job. Moreover, there are only a few gold-standard AFPs that can be employed towards the design of reliable classification models.

Effective classification of the AFPs is of prime importance, however, predicting them manually requires extensive labor and time. Considering the advancements in the computational methods and rapid development in machine learning-based solutions, many researchers have proposed artificial intelligence-based algorithms for a variety of protein classification problems \cite{park2020e3, khan2015machine, barman2014prediction}. For example, in \cite{park2020e3} \textit{Park et al.} proposed a deep learning-based latent space classification model for E3-target protein pairs.  In \cite{khan2015machine}, \textit{Khan et al.} proposed two classification approaches for AFPs and extracellular matrix proteins (ECMPs). In \cite{barman2014prediction}, the prediction of the interactions between the viral and host proteins was performed with the help of supervised machine learning methods. 

Machine learning frameworks include two elemental portions; the feature extraction and the classification. It is necessary for the training of a machine learning algorithm that distinguishing features are derived from the dataset. Accordingly, many feature extraction methods are utilized, and sometimes the obtained features are further filtered via feature selection methods to obtain the most relevant features. There is a variety of classifiers available including neural networks, decision trees, and nearest neighbors and the selection of which is dependent on the nature of the application. The computational methods for the prediction of the AFPs have been discussed by \textit{Fang et al.} in \cite{wang2019brief}. The machine learning-based solution to the diversified problem of the AFPs classification was first proposed by \textit{Kandaswamy et al.} as AFP-Pred \cite{kandaswamy}. The sequences of the AFPs were encoded and a resultant feature vector containing 119 attributes was obtained. Out of those attributes, the dominant features were selected by applying \emph{ReliefF}, and the \emph{random forest (RF)} classifier was trained on them to perform the classification. In \cite{yu2011identification}, n-peptide compositions and physicochemical features were extracted from the protein sequences. The dominant features were selected using a \emph{genetic algorithm} (GA) and the resultant features were used to train a \emph{support vector machine} (SVM) based classifier. The \emph{position-specific scoring matrix}  (PSSM) profiles, representing the evolutionary information in the sequences, were utilized in \cite{xiaowei}.
The prediction model of \cite{xiaowei} was named \emph{AFP\_PSSM} and it is based on the SVM classifier. In \cite{mondal2014chou}, the pseudo amino acid compositions (\emph{pseAAC}) were used for the formulation of the feature set. The features were utilized to train the SVM-based classifier, coined as \emph{AFP-PseAAC}. Another method named as \emph{afpCOOL} \cite{eslami2018afpcool} was proposed to classify the AFPs by utilizing four descriptors as a feature vector. The feature vector composed of hydropathy, physicochemical properties, amino acid composition, and evolutionary profile was used to train the SVM-based classifier.

Current machine learning methods provide reasonably good classification models, however, they do not provide a sample-specific relationship which is important for sub-class/type prediction. Towards this end, we propose a sample-specific classification method in this research using a sparse reconstruction classification method. Specifically, a linear model and an \emph{over-complete dictionary matrix} (ODM) of known AFPs are designed to predict a sparse class-label vector that provides a sample-association score.  Later, we reconstruct two pseudo-samples through the delta-rule using lower and upper parts of the sample-association vector and assign class labels based on the minimum recovery score. The detailed method is explained in Section \ref{sec:method}, while results and conclusions are provided in Section \ref{sec:result} and Section \ref{sec:conclusion}, respectively.

\section{Proposed Approach}\label{sec:method}

\subsection{Dataset}
\begin{figure}[!htb]
\centering
\includegraphics*[width=6cm]{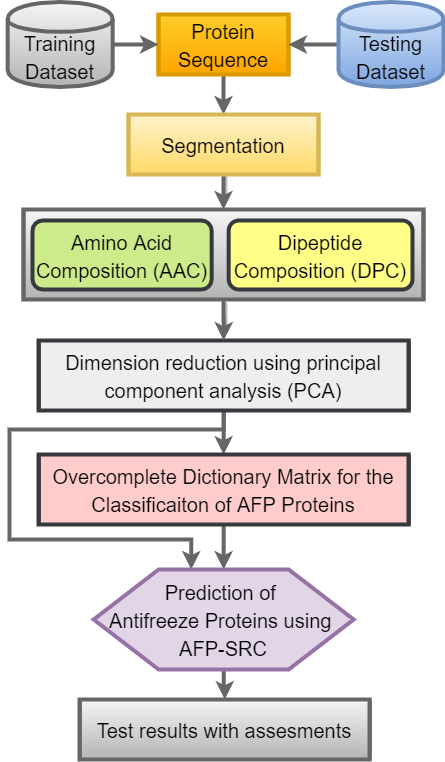}
\caption{Flow chart of the proposed model.}
\label{fig:flow}
\end{figure}

The standard dataset is obtained from AFP-Pred \cite{kandaswamy}. The dataset was initially derived from \emph{Pfam database seed} containing 221 AFPs. For the removal of redundancy among the sequence, the PSI-BLAST program was implied with a strict threshold $(E=0.001)$. After a manual check, the sequence identity was decreased to up to $40\%$ using the CD-HIT program. The final dataset contained a total of $481$ AFPs and $9493$ non-AFPs. For training and testing, the dataset was split by randomly selecting $600$ samples, i.e., $300$ AFPs and $300$ non-AFPs for training, while the remaining $181$ AFPs and $9193$ non-AFPs were used for test purpose. The design flow of the proposed method is illustrated in Fig. \ref{fig:flow}.

\subsection{Amino Acid Sequence Encoding}

\begin{figure}[!htb]
\centering
\includegraphics*[width=8cm]{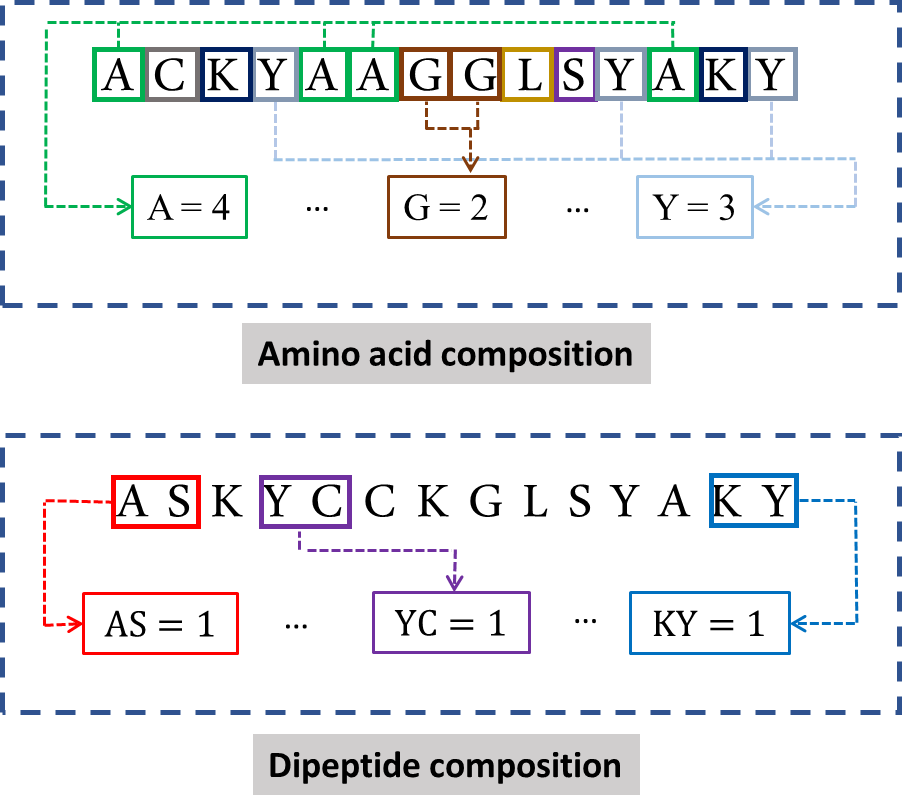}
\caption{Amino acid and Dipetide composition illustration}
\label{fig:AAC}
\end{figure}

Protein sequences are generally stored in FASTA format where $20$ alphabets are used to represent $20$ essential amino acids. Besides, each protein sequence has a particular length which makes it even more difficult to establish a mathematical link between the sequence and its class label. Since machine learning algorithms require numeric representation to establish a connection between the class label and the input sample, it is necessary to encode peptide sequence into a format where maximum information can be represented into a fixed size numerical format. Many encoding schemes have been suggested to be suitable for the numerical representation of AFPs \cite{8941932, usman2020afp}. However, the two popular and simple methods to encode peptide sequences are \emph{amino-acid composition} (AAC), and \emph{di-peptide composition feature} (DPC) \cite{khan2016rafp, Affan2020}.  The AAC represents the frequency of $20$ essential amino acids in a sequence, therefore, it generates a fixed-sized feature vector for each protein. Similarly, DPC is a frequency of the second-order permutation of amino-acids i.e., it is a frequency vector of the pair of amino-acids, therefore, it generates a feature vector of size $20\times20 = 400$. The AAC and DPC, illustrated in Fig. \ref{fig:AAC},  are found to be robust feature encoding schemes, yet, they can only extract the global features of the proteins. On the other hand, many functions of the proteins are associated with the localized domains of peptides in the protein. Similar to short-time Fourier transform which helps in finding the frequency components of the localized region of the signal, the segmented ACC and DPC can provide localized features. Accordingly, each protein is divided into two equal segments in this study and ACC+DPC features are extracted from each segment resulting in $420\times2=840$ features. 

\subsection{Dimension Reduction Using PCA}

\begin{figure}[!htb]
\centering
\includegraphics*[width=8cm]{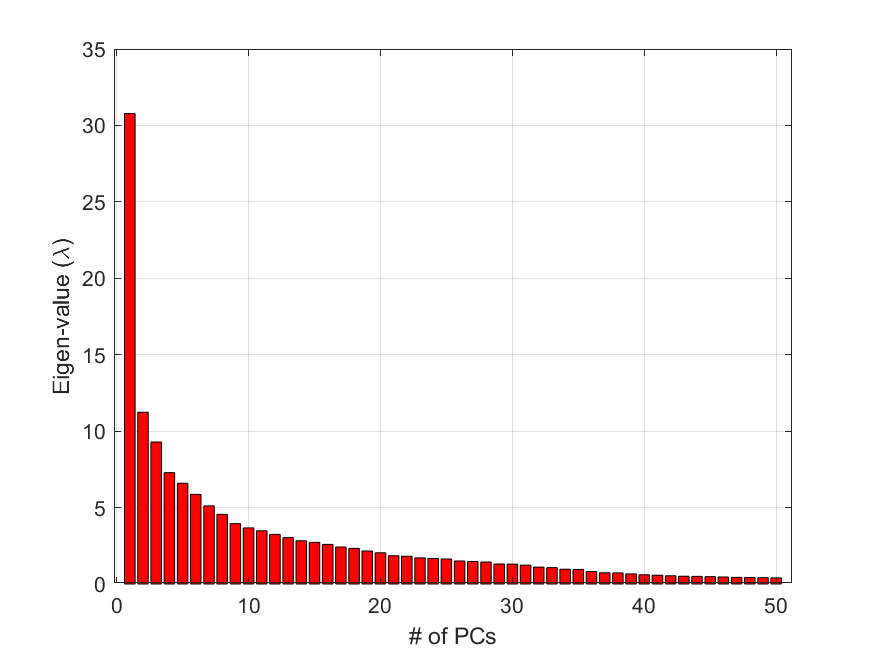}
\caption{Eigenvalues of the top $50$ principal components.}
\label{fig:PCA}
\end{figure}

To design a robust classification model, it is important to incorporate only the most useful features that provide maximum information about the data. Two popular techniques in use are feature selection and dimension reduction. In the feature selection approach, useful attributes are found by filtration or wrapping method. While in dimension reduction all the attributes are first transformed into a compressed form,  such as principal component or kernel representation. The \emph{principal component analysis} (PCA) is a widely used method to extract a noise-free representation of data in a reduced dimension space \cite{naseem2017ecmsrc}. In this study, PCA is used to transform $840$ features into $840$ components. The $840$ components are obtained using the covariance matrix of the training dataset. The vectors are arranged in descending order depending on their significance which is defined by their eigenvalues. Figure ~\ref{fig:PCA} shows the eigenvalues of top $50$ vectors.

\subsection{Over-complete Dictionary Matrix for Classifying AFPs}

After encoding and feature compressing, an ODM is formed using training samples. The ODM is used for sparse representation classification in which each AFP, as well as each non-AFP, is characterize using class index $c$. Specifically, $c=1$ and $c=2$ correspond to AFPs and non-AFPs, respectively.

Let $S$ be the number of training samples from each class  such that $\mathbf{v}_i^{(c)}$ $\in\mathbb{R}^{p}$ represents the $i^{th}$ training sample from $c^{th}$ class, for $c=1,2$ and $p$ the number of PCAs.  Then, an ODM, $\mathbf{T}\in \mathbb{R}^{p \times 2S}$, is formed by concatenating all training samples as  
\begin{align}
\mathbf{T}:=\begin{bmatrix}
\mathbf{v}_1^{(1)} & \mathbf{v}_2^{(1)}& \cdots & \mathbf{v}_S^{(1)} & \mathbf{v}_1^{(2)} & \mathbf{v}_2^{(2)} & \cdots & \mathbf{v}_S^{(2)}\end{bmatrix}.
\end{align} 
A test sample $\mathbf{t} \in \mathbb{R}^{p}$ can be represented as
\begin{align}
\mathbf{t}=\mathbf{T}\bomega, 
\end{align}
where the coefficient vector $\bomega\in\mathbb{R}^{2S}$ is defined by
\begin{align}
\label{opt1}
\bomega=\begin{bmatrix} \omega_{1}^{(1)} & \omega_{2}^{(1)}& \ldots & \omega_{S}^{(1)}& \omega_{1}^{(2)}& \omega_{2}^{(2)}& \ldots &\omega_{S}^{(2)}\end{bmatrix}^T.
\end{align}

If true class of a test sample $\mathbf{t}$  is the $c^{th}$ class, all entries of $\bomega$ should be zero except $\omega_{1}^{(c)},\omega_{2}^{(c)},\cdots, \omega_{S}^{(c)}$.  According to sparse reconstruction theory, if dictionary matrix $\mathbf{T}$ is given, the sparse vector $\bomega$ can be recovered \cite{naseem2008sparse, naseem2010sparse}.  In principle, the sparsest $\bomega$ can be sought as the solution to the optimization problem
\begin{equation}
\arg \ds{\min_\bomega} \left\|\bomega\right\|_{0}\ \ \mbox{subject to}\ \  \mathbf{t}=\mathbf{T}\bomega,
\label{l0}
\end{equation}
where $\left\|\cdot\right\|_{0}$ is the $l_0$-norm (counting the number of non-zeros entries in the vector). 

The constrained optimization problem \eqref{l0} is non-convex which makes it hard to find the optimal vector $\bomega$.  Several algorithms have been proposed in the literature to recover the sparse vector $\bomega$ by solving a convex relaxtion of the constrained optimization problem \eqref{l0}.  The \emph{basis pursuit} (BP) algorithm, for instance, is one of those algorithms that makes use of the $l_1$-norm to solve the \emph{relaxed} optimization problem \cite{sparse7}
\begin{equation}
\arg{\min_\bomega} \left\|\bomega\right\|_{1}\ \ \mbox{subject to}\ \  \mathbf{t}=\mathbf{T}\bomega.
\label{l1}
\end{equation}
Under certain conditions on the isometry constant of the matrix $\mathbf{T}$, the sparse vector $\bomega$ can be ecovered with high probability by solving \eqref{l1} using the BP algorithm (see, for instance, \cite{sparse2}, \cite{sparse3}).  

Notice that, $\bomega$ is expected to have high-value entries corresponding to the columns of $\mathbf{T}$ that are relevant to the class label of the probe $\mathbf{t}$.  This embedded information about the class label of $\mathbf{t}$ can be used to identify $\mathbf{t}$. Let 
\begin{align}
r_c(\mathbf{t})=\left\|\mathbf{t}-\mathbf{T}\delta_c(\bomega)\right\|_2, \qquad \ c=1,2
\end{align}
where the vector $\delta_c$ has all zero entries except at the locations corresponding to class $c$ where the value is one.  The decision is ruled in favor of the class with the minimum reconstruction error, i.e., 
\begin{align}
\mbox{class label}(\mathbf{t})=\arg {\min_c} \left( r_c (\mathbf{y})\right).
\end{align}

The MATLAB implementation of the proposed method is available at the author's GitHub page (\href{https://github.com/Shujaat123/AFP-SRC}{https://github.com/Shujaat123/AFP-SRC}).

\section{Experimental Results} \label{sec:result}

The proposed algorithm was evaluated for true positive rate (sensitivity), true negative rate (specificity), prediction accuracy,  Matthew's Correlation Coefficient (MCC), balanced accuracy and  Youden's index with the following definitions:
\begin{align}
&{\rm Sensitivity} = \frac{TP}{TP+FN}, 
\\
&{\rm Specificity} = \frac{TN}{TN+FP},
\\
&{\rm Accuracy} = \frac{TP+TN}{TP+TN+FP+FN},
\\
&{\rm  MCC} ={\frac{TPTN-FPFN}{\sqrt{\Delta}}},
\\
& \Delta = (TP+FP)(TN+FN)(TP+FN)(TN+FP),
\\
&\text{Balanced Accuracy} = \frac{{\rm Sensitivity} + {\rm Specificity}}{2},
\\
&\text{Youden's  Index} = {\rm Sensitivity} + {\rm Specificity} - 1,
\\
&\text{F$1$ Score} = 2* \frac {{\rm Precision} *{\rm Recall}}{{\rm Precision} + {\rm Recall}}.
\end{align}

The true positive (TP) indicates the correctly classified positive proteins and the true negative (TN) indicates the correct classification of proteins from the negative class. The false positive (FP) (resp. false-negative (FN)) represents the incorrect predictions of the positive (resp. negative) class proteins. The range of MCC lies between the values $-1$ and $1$, respectively indicating the worst and the best classification of the classifier. For class-specific measures, balanced accuracy and Youden's index are implied and F-score is calculated to obtain the harmonic mean of the precision and recall, representing the efficacy of the classifier. 

The performance of the proposed classifier was evaluated by incremental variation in the number of principal components. In particular, nineteen different feature sets were tested consisting of $\{10, 20, \cdots, 100, 150, 175, \cdots 250, 300, 400, \cdots, 600\}$ principal components, and all the above mentioned statistical measures were evaluated.

\subsection{Evaluation of Robustness of Dictionary Matrix}

\begin{figure}[!htb]
\centering
\includegraphics*[width=8cm]{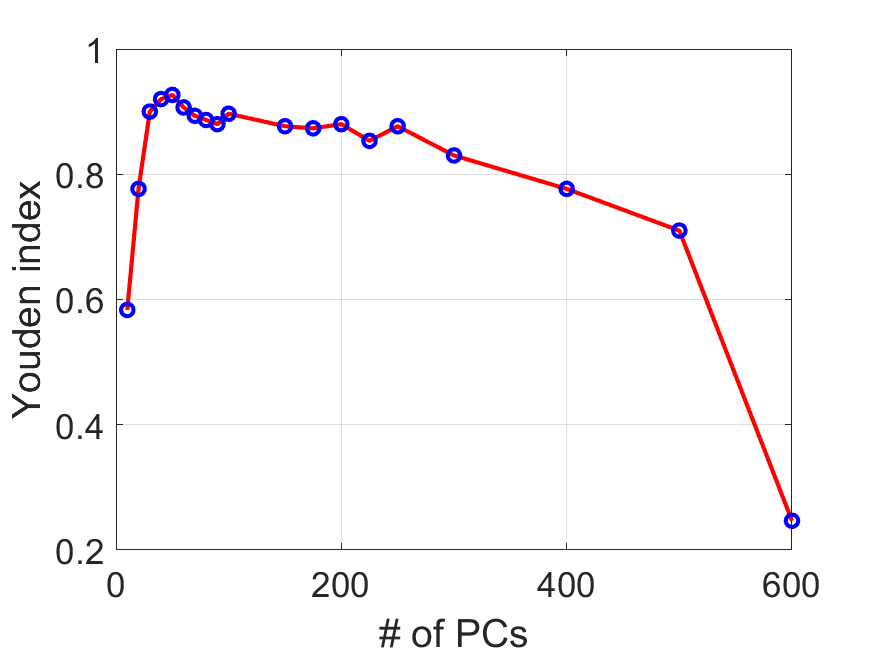}
\caption{Youden's-Index of AFP-SRC on training dataset for different number of PCAs.}
\label{fig:youden_train}
\end{figure}

Before evaluating the performance on the test dataset, we first evaluate the robustness of the dictionary matrix. Towards this end, we first normalize the training samples and add the Gaussian noise of unit variance and zero mean in the dictionary. The robustness of the dictionary is measured in the form of performance statistics defined above and results are reported in Table \ref{tbl:ablation_train}. To summarize the findings, the Youden's index metric for the training dataset is plotted in Fig.~\ref{fig:youden_train}. It can be seen that the method has some tolerance against noise, which means that the SRC can recover the true sample class even with the noisy dictionary. An important point to notice is that the PCA provides filtration by separating the useful information from noise with the help of singular value decomposition (SVD). Through SVD, the correlated signal appears in top eigenvectors while uncorrelated (noise) components appear in lower eigenvectors. Therefore, with the increasing number of principal components, the performance of SRC decreases.
\begin{table*}[!htb] 
\caption{Performance statistics of AFP-SRC using training dataset on different number of principal components (PCs).}
\label{tbl:ablation_train}
\begin{center}

\resizebox{0.9\textwidth}{!}{
\begin{tabular}{cccccccc}
\hline
\textbf{PCs} & \textbf{Youden's Index} & \textbf{Balanced Accuracy} & \textbf{MCC} & \textbf{Sensitivity} & \textbf{Specificity} & \textbf{Accuracry} & \textbf{F1-Score} \\ \hline
10  & 0.35 & 67.50& 0.37 & 86.33 & 48.67 & 67.50  & 0.72 \\ \hline
20  & 0.70 & 85.33 & 0.71 & 92.00 & 78.67 & 85.33 & 0.86 \\ \hline
30  & 0.89 & 94.66 & 0.89 & 97.66 & 91.67 & 94.67 & 0.94 \\ \hline
40  & 0.95 & 97.66 & 0.95 & 99.33 & 96.00 & 97.67 & 0.97  \\ \hline
50  & 0.95 & 97.83 & 0.95 & 100.00 & 95.67 & 97.83 & 0.98 \\ \hline
60  & 0.96 & 98.33 & 0.96 & 100.00 & 96.67 & 98.33 & 0.96 \\ \hline
70  & 0.94 & 97.33 & 0.94 & 98.67 & 96.00 & 97.33 & 0.97 \\ \hline
80  & 0.95 & 97.66 & 0.95 & 99.00 & 96.33 & 97.67 & 0.97  \\ \hline
90  & 0.95 & 97.66 & 0.95 & 98.67 & 96.67 & 97.67 & 0.97 \\ \hline
100 & 0.96 & 98.16 & 0.96 & 99.67 & 96.67 & 98.17 & 0.98 \\ \hline
150 & 0.95 & 97.83 & 0.95 & 99.00 & 96.67 & 97.83 & 0.97 \\ \hline
175 & 0.96 & 98.33 & 0.96 & 99.33 & 97.33 & 98.33 & 0.98 \\ \hline
200 & 0.97 & 98.83 & 0.97 & 100.00& 97.67 & 98.83 & 0.98 \\ \hline
225 & 0.97 & 98.66 & 0.97 & 99.33 & 98.00 & 98.67 & 0.98 \\ \hline
250 & 0.96 & 98.16 & 0.96 & 99.33 & 97.00 & 98.17 & 0.98 \\ \hline
300 & 0.99 & 99.50 & 0.99 & 99.67 & 99.33 & 99.50 & 0.99  \\ \hline
400 & 0.98 & 99.33 & 0.98 & 99.67 & 99.00 & 99.33 & 0.99 \\ \hline
500 & 0.97 & 98.83 & 0.97 & 99.67 & 98.00 & 98.83 & 0.98 \\ \hline
600 & 0.93 & 96.83 & 0.93 & 98.00 & 95.67 & 96.83 & 0.96 \\ \hline
\end{tabular}}
\end{center}
\end{table*}

\subsection{Evaluation of classification performance.}

\begin{figure}[!htb]
\centering
\includegraphics*[width=8cm]{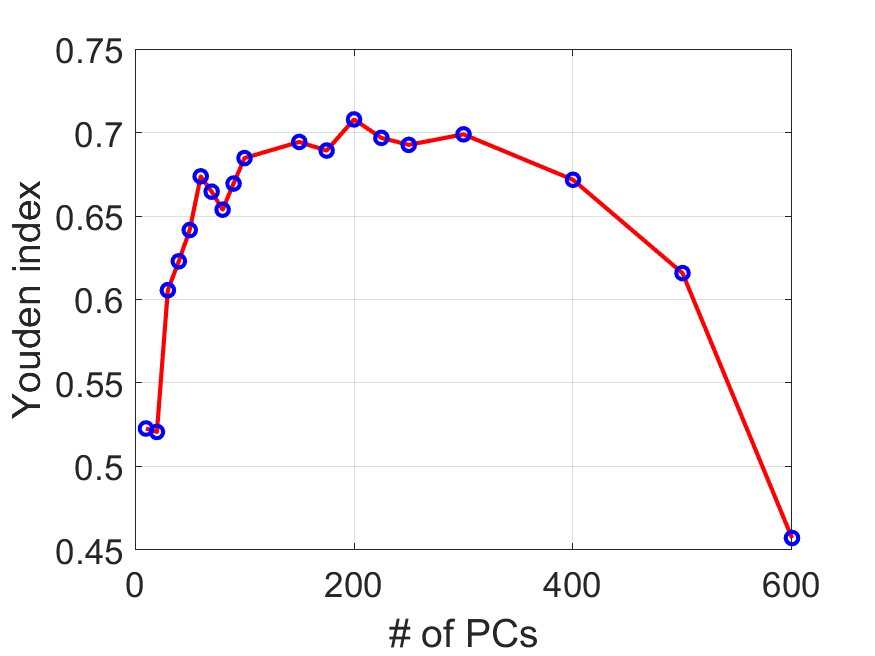}
\caption{Youden's-Index performance of AFP-SRC on test dataset for different number of PCAs.}
\label{fig:youden_test}
\end{figure}

For test samples, as depicted in Table \ref{tbl:ablation_test}, a gradual improvement in the performance of the classifier is observed when the number of principal components is increased. In Fig.~\ref{fig:youden_test}, the same can be seen graphically in the form of Youden's index measured for a different number of principal components. The best statistics were obtained with $200$ principal components and with further increase in the principal components, the performance begins to decline. This indicates that the $200$ principal components are sufficient for the classification of AFPs. Moreover, the principal components with lower eigenvalues are not useful and most likely represent the noise. Therefore, the data is projected on the top $200$ components for the proposed algorithm. 

\begin{table*}[!htb] 
\caption{Performance statistics of AFP-SRC using test dataset on different number of principal components (PCs).}
\label{tbl:ablation_test}
\begin{center}
\resizebox{0.9\textwidth}{!}{
\begin{tabular}{cccccccc}
\hline
\textbf{PCs} & \textbf{Youden's Index} & \textbf{Balanced Accuracy} & \textbf{MCC} & \textbf{Sensitivity} & \textbf{Specificity} & \textbf{Accuracry} & \textbf{F1-Score} \\ \hline
10  & 0.52 & 76.13 & 0.14 & 88.39 & 63.87 & 64.35 & 0.08 \\ \hline
20  & 0.52 & 76.03 & 0.15 & 85.63 & 66.43 & 66.80 & 0.09 \\ \hline
30  & 0.60 & 80.27 & 0.18 & 87.84 & 72.71 & 73.47   & 0.11 \\ \hline
40  & 0.62 & 81.14 & 0.22   & 85.08 & 77.20 & 77.35 & 0.12 \\ \hline
50  & 0.64 & 82.08 & 0.21  & 85.08 & 79.07 & 79.19 & 0.13 \\ \hline
60  & 0.67 & 83.68 & 0.22 & 88.95 & 78.42 & 78.62 & 0.13 \\ \hline
70  & 0.66 & 83.22 & 0.21 & 87.29 & 79.16 & 79.32 & 0.13 \\ \hline
80  & 0.65 & 82.69   & 0.21 & 87.29 & 78.08 & 78.26 & 0.13 \\ \hline
90  & 0.66 & 83.47 & 0.21 & 88.39 & 78.54 & 78.73 & 0.13 \\ \hline
100 & 0.68 & 84.23 & 0.22 & 90.05 & 78.41 & 78.63 & 0.13 \\ \hline
150 & 0.69 & 84.71 & 0.22 & 90.60 & 78.82 & 79.05 & 0.14 \\ \hline
175 & 0.68 & 84.45 & 0.22 & 89.50 & 79.41 & 79.60 & 0.14 \\ \hline
\textbf{200} & \textbf{0.71} & \textbf{85.40} & \textbf{0.23} & \textbf{91.16} & \textbf{79.62} & \textbf{79.84} & \textbf{0.14}  \\ \hline
225 & 0.69 & 84.83 & 0.23 & 89.50 & 80.17 & 80.35 & 0.14 \\ \hline
250 & 0.69 & 84.62 & 0.23 & 88.39 & 80.86 & 81.73   & 0.15 \\ \hline
300 & 0.69 & 84.94 & 0.24 & 88.39 & 81.49 & 81.62 & 0.15  \\ \hline
400 & 0.67 & 83.58 & 0.22 & 86.18 & 80.98 & 81.08 & 0.14 \\ \hline
500 & 0.61 & 80.79 & 0.20 & 82.87 & 78.70 & 78.78 & 0.13 \\ \hline
600 & 0.45 & 72.86 & 0.13 & 74.58 & 71.13 & 71.20 & 0.09 \\ \hline
\end{tabular}}
\end{center}
\end{table*}

\begin{table*}[!ht] 
\caption{Performance comparison of AFP-SRC and contemporary methods on test dataset.}
\label{tbl:comparison}
	\begin{center} 
		\resizebox{0.9\textwidth}{!}
		{\begin{tabular}{lcccccccc}
			\hline
			{\bf Methods} & {\bf Youden's Index} & {\bf Sensitivity} &  {\bf Specificity} & {\bf Accuracy}  & {\bf Balanced Accuracy} &{\bf Classifier} \\
			\hline
			iAFP \cite{yu2011identification} & 0.10  &  13.2\% & 97.0\%  &  95.3\%   &  55.1\% & SVM \\ 
			\hline
			AFP-Pred \cite{kandaswamy} & 0.63 &  84.6\%  &  82.3\% &83.3\%  & 83.4\% & RF\\
			\hline
			AFP\_PSSM \cite{xiaowei} &  0.69 &   75.8\% &  93.2\%  & 93.0\%  & 84.5\% & SVM\\
			\hline
			afpCOOL \cite{eslami2018afpcool} & 0.70 &  72.0\% & 98.0\% &  96.0\% & 85.0\%  & SVM\\
			\hline
			AFP-PseAAC \cite{mondal2014chou} & 0.70 &86.1\% & 84.7\%  &84.7\% & 85.4\% & SVM\\
			\hline
			AFP-SRC & 0.71 &  91.1\% & 79.6\% &  79.8\% & 85.4\%  & SRC\\
			\hline
		\end{tabular} }
	\end{center}
\end{table*}

Table \ref{tbl:comparison}, shows the performance of the proposed AFP-SRC method with the existing methods. The number of principal components in the algorithm is chosen to be $200$. It is noteworthy to point out that in the proposed AFP-SRC there is no training phase and the training data is only used to generate a dictionary matrix. However, for a fair comparison, the training and testing of all the methods are done by keeping a similar configuration of the dataset. The accuracy parameter indicates the overall accuracy of the classifier, which can be deceiving in the case of imbalanced training and testing data. Therefore, we emphasize on the class-specific evaluation parameters. The proposed method outperforms the existing methods in terms of the class-specific evaluation parameters, i.e., Youden's index and balanced accuracy. The method yields the best sensitivity results which substantiate its ability to effectively project the features of the AFPs. In particular, AFP-SRC achieved highest Youden's index value of $0.71$ which is $61\%$,$8\%$,$2\%$,$1\%$, and $1\%$ higher than the iAFP \cite{yu2011identification}, AFP-Pred \cite{kandaswamy}, AFP-PSSM \cite{xiaowei}, AFP-PseAAC \cite{mondal2014chou}, and afpCOOL \cite{eslami2018afpcool} respectively. This suggests that the proposed method may serve as a platform for the designing of novel AFPs or AFP like proteins. This effectiveness is also reflected in the high Youden's index value indicating the distinguishing potential between AFPs and non-AFPs. Likewise, the balanced accuracy achieved by the proposed method is also comparable. In particular, AFP-SRC achieved highest balanced accuracy value of $85.4\%$ which is $30.3\%$,$2\%$,$0.9\%$, and $0.4\%$ higher than the iAFP \cite{yu2011identification}, AFP-Pred \cite{kandaswamy}, AFP-PSSM \cite{xiaowei} and afpCOOL \cite{eslami2018afpcool} respectively, and equals to AFP-PseAAC \cite{mondal2014chou}.

\section{Conclusion}\label{sec:conclusion}

Antifreeze proteins are essential for the cold-adapted organisms since it prevents the body fluids from freezing and are commonly used in medical and food industry in a variety of applications. The sequence and structural diversity in the antifreeze proteins make their classification a challenging task. We present a computational approach, coined as AFP-SRC, to effectively classify the AFPs from non-AFPs based on the sample specific classification method using sparse representation. A sparse class-label vector is predicted using an over-complete dictionary of known samples and a sample-association score is obtained. Class labels are assigned using the minimum recovery score via delta-rule. The proposed method is evaluated for the well-known statistical parameters and is found to outperform the existing methods. The results indicate higher sensitivity of the proposed AFP-SRC method which could be useful in the understanding of the structural and chemical properties and development of novel AFPs. 
	
	\bibliographystyle{IEEEtran}
	\bibliography{main}

\end{document}